\documentclass{article}

\usepackage[preprint]{corl_2026} 

\usepackage{graphicx}
\usepackage{booktabs}
\usepackage{hyperref}
\usepackage{adjustbox}
\usepackage{makecell}
\usepackage{subcaption}
\usepackage{amsmath}
\usepackage{amssymb}
\usepackage{xcolor}

\definecolor{darkgreen}{rgb}{0, 0.75, 0}

\newcommand{\para}[1]{\noindent\textbf{#1}}
\newcommand{\ours}{DynaMOMA}

\definecolor{mygreen}{HTML}{75F6A9}
\definecolor{myred}{HTML}{FF7E60}

\title{DynaMOMA: Instantaneous Prediction of Grasp Poses for Mobile Manipulation of Dynamic Objects}

\author{
    Zhinan Yu$^{*, 1}$, Junyan Xu$^{*, 1}$, Jiazhao Zhang$^{2}$, Zheng Qin$^{3}$, Yijie Tang$^{1}$, Yuhang Huang$^{1}$,\\ \textbf{Yihan Cao}$^{1}$\textbf{,} \textbf{Zhiyuan Yu}$^{4}$\textbf{,} \textbf{Yongjun Wang}$^{1}$\textbf{,} \textbf{Renjiao Yi}$^{1}$\textbf{,} \textbf{Chenyang Zhu}$^{1}$\textbf{,} \textbf{Kai Xu}$^{5,\dagger}$ \\
    $^{1}$ National University of Defense Technology \quad
    $^{2}$ CFCS, Peking University \\
    $^{3}$ Defense Innovation Institute, Academy of Military Sciences \\
    $^{4}$ Wuhan University \quad
    $^{5}$ Institute of AI for Industries, Chinese Academy of Sciences
}

\begin{document}
\maketitle

\begingroup
  \renewcommand{\thefootnote}{} 
  \footnotetext{$^{*}$ Equal Contribution. \quad $^{\dagger}$ Corresponding Authors.}
\endgroup

\vspace{-20pt}
\begin{abstract}
Mobile manipulation is a fundamental robotics task and has advanced rapidly in recent years, enabling robots to navigate, reach, and interact with objects in complex environments.
However, mobile manipulation of dynamic objects remains highly challenging, as robots must coordinate the mobile base and arm while adapting to continuously evolving target poses.
A key challenge lies in predicting temporally consistent short-horizon grasp trajectories from dynamic observations.
In this work, we propose \ours{}, a dynamic mobile manipulation framework that couples instantaneous grasp trajectory prediction with whole-body control policy.
Our predictor uses an anchor-based diffusion model to generate temporally consistent short-horizon grasp trajectories conditioned on historical observations.
The predicted trajectories are then encoded as compact features and fed to a whole-body reinforcement learning policy, which controls the mobile manipulator for dynamic grasping.
We further introduce a anticipation-guided reward that equips the policy with an anticipatory grasping horizon by adaptively shifting the target from the current grasp observation to the instantaneously predicted grasp trajectory.
Through extensive experiments in Isaac Gym simulation, we show that our method achieves strong performance in mobile manipulation of dynamic objects across diverse settings and grasping metrics.
Furthermore, our predictor and policy demonstrate strong generalizability in real-world experiments.

\end{abstract}



\section{Introduction}
\label{sec:intro}

\begin{figure}[t]
\centering
\includegraphics[trim=0cm 0cm 0cm 0cm, clip, width=\linewidth]{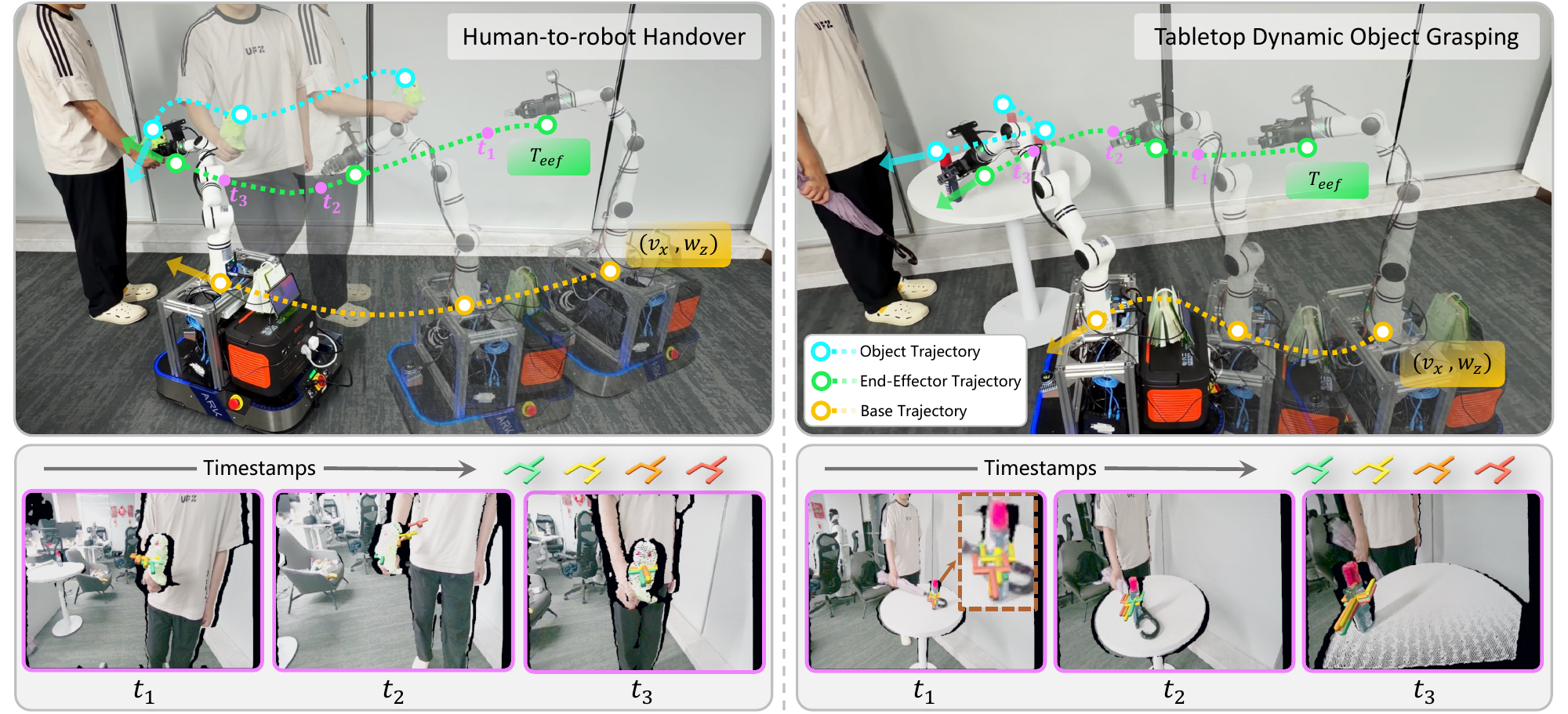}
\caption{
\textbf{Illustration of \ours{} in real-world mobile manipulation tasks.} \textbf{Top row:} Third-person views of the human-to-robot handover task and the tabletop dynamic object grasping task. \textbf{Bottom row:} Chronological first-person point clouds captureed by the wrist-mounted camera at different timestamps ($t_{1}$, $t_{2}$, $t_{3}$). The instantaneous prediction of grasp poses are visualized as sequential grippers, color-coded from \textcolor{mygreen}{green} to \textcolor{myred}{red} to represent the temporal progression.
}
\vspace{-10pt}
\label{fig:teaser}
\end{figure}

Mobile manipulation has emerged as a core capability in robotics, with applications ranging from household assistance and warehouse logistics to manufacturing~\citep{brock2016mobility, hebert2015mobile, wang2025trackvla, watkins2022mobile, kalashnikov2018scalable, mahler2019learning, li2026llm}.
Among its open challenges, grasping a dynamically moving object stands out as one of the most demanding.
The robot must coordinate its mobile base and articulated arm to manipulate a target object whose pose changes continuously.
This requires the system not only to react to the current observation, but also to anticipate where the object will be in the near future. 

Existing approaches to mobile manipulation broadly fall into three categories.
Classical planners rely on 3D reconstruction and geometric reasoning, which limits their real-time performance in dynamic settings~\citep{sun2022motion, chen2023hierarchical, patki2020language, burgess2024reactive, wu2024real}.
End-to-end reinforcement learning (RL) policies improve efficiency by mapping observations directly to actions, yet they can only respond to what is currently observed and therefore often lag behind moving targets~\citep{yokoyama2023asc, jauhri2022robot, wang2020learning, sun2022fully, zhang2024gamma, wang2025quadwbg}.
In contrast, explicit forecasting methods estimate object motion to reduce perception noise, but they do not predict grasp poses~\citep{mulling2013tabletennis, kim2014catching, dambrosio2024tabletennis}.
This distinction is critical because a grasp pose is a geometry-conditioned action target rather than an object state. In dynamic mobile grasping, active camera motion can rapidly change both visual observations and camera-object relative poses. As a result, extrapolating the object's motion alone is insufficient to determine a reliable grasping trajectory.

To address these issues, we propose \ours{}, a dynamic mobile grasping framework that couples instantaneous grasp trajectory prediction with whole-body control policy.
Our key insight is that short-horizon grasp predictions can guide the policy beyond the current observation, transforming reactive pursuit of moving objects into feedforward control.
Specifically, we train an anchor-based diffusion predictor that generates temporally consistent, multi-mode\footnote{Following DiffusionDrive~\citep{liao2025diffusiondrive}, ``multi-mode'' in this paper refers to diverse grasp trajectories.} instantaneous grasp trajectories~\citep{liao2025diffusiondrive}.
Conditioned on historical observations, the predictor maintains consistency with prior predictions while capturing the evolving grasp feasibility.
We further train a whole-body RL policy with a anticipation-guided reward that adaptively balances current observations and instantaneous predictions, enabling stable approach from afar and smooth feedforward control near the grasp window.
To support training and evaluation, we build a dynamic grasping environment in Isaac Gym~\citep{makoviychuk2021isaac} with YCB objects~\citep{calli2015ycb}, from which we collect $5$K training sequences and design a dynamic object mobile manipulation (DOMM) benchmark.
To further bridge the Sim-to-Real gap, we compile an additional $1.2$K real-world sequences by combining data collected from our physical platform and extracted from the DexYCB dataset~\citep{chao2021dexycb}.

We conduct extensive experiments in both simulation and real-world settings.
On the DOMM benchmark, \ours{} significantly outperforms reactive and prediction-augmented baselines, and real-world deployment further confirms robust dynamic grasping under diverse object motions.
In summary, we propose a novel grasp trajectory predictor to enable instantaneous prediction of grasp poses, and an anticipation-guided reward that steers the whole-body control policy toward anticipated grasp trajectory. Futhermore, we build a dynamic mobile grasping system that demonstrates strong performance in both simulation and real-world deployment.


\section{Related Work}
\label{sec:related_work}

\para{Mobile Manipulation.}
Mobile manipulation requires coordinated control of a mobile base and an articulated arm to reach and grasp objects in unstructured environments.
Classical methods typically formulate this problem as joint motion planning or trajectory optimization over all degrees of freedom~\citep{minniti2019whole, sleiman2021unified, jiao2021consolidating}. However, repeatedly re-solving such optimization problems is computationally expensive and often struggles with fast-moving targets.
To improve responsiveness, recent work has turned to reactive reinforcement learning policies that map observations directly to whole-body commands~\citep{hu2023causal, fu2023deep}.
For example, GAMMA~\citep{zhang2024gamma} fuses raw grasp candidates across past frames to produce coordinated base--arm actions, while VBC~\citep{liu2024vbc} maps egocentric RGB observations directly to joint commands for legged loco-manipulation.
Despite their reactivity, these policies still condition primarily on the current observation.
Since the object may continue to move, the grasp poses estimated from the current frame can quickly become outdated, causing the robot to chase the object reactively rather than move to intercept it.

\para{Dynamic Object Interception.}
A complementary line of work enables robots to intercept moving targets through explicit motion forecasting.
In robotic ball sports, for instance, extended Kalman filters or ballistic models are used to predict where a ball will arrive, allowing the robot to pre-position its end-effector~\citep{mulling2013learning}.
Such forecasting-based strategies have achieved strong results in table tennis~\citep{dambrosio2025achieving}, ball catching~\citep{kim2014catching}, and batting~\citep{jia2019batting}.
Beyond ball sports, related methods for dynamic grasping on conveyors~\citep{akinola2021dynamic} and human handovers~\citep{yang2021reactive, zhang2023flexible} also commonly rely on specific kinematics assumptions.
However, these methods can degrade when object motion is unpredictable or contact-dominated, as often occurs in general mobile manipulation where a person may change direction or pass an object along a curved path.
Our work preserves the anticipatory benefit of forecasting while replacing the kinematics models with a learned predictor that does not assume a particular form of the underlying dynamics.

\para{Diffusion Models for Trajectory Prediction.}
Denoising diffusion models have recently emerged as a powerful tool for robotic action and trajectory generation.
Diffusion Policy~\citep{chi2025diffusion} generates multi-step action sequences by iteratively denoising Gaussian noise, demonstrating strong expressiveness on contact-rich manipulation tasks.
Subsequent work has extended this paradigm to 3D manipulation~\citep{ze20243d}, locomotion~\citep{huang2024diffuseloco}, and long-horizon planning~\citep{janner2022planning, huang2025ladi}.
A practical limitation, however, is that recovering structured outputs from isotropic noise often requires a long denoising process.
Anchor-based diffusion methods alleviate this issue by initializing generation from learned anchor modes rather than pure noise, allowing each branch to refine within a local neighborhood and reducing the number of required denoising steps~\citep{liao2025diffusiondrive,wang2025trackvla,hoeg2024streaming}.
Our predictor builds on this anchor-based formulation, but focuses on forecasting grasp trajectories rather than generating ego-actions or vehicle trajectories.
We further couple the predictor with a whole-body control policy through an anticipation-guided reward, thereby bridging prediction and control.


\section{Method}
\label{sec:method}
\subsection{Preliminary}

\para{Task formulation.}
Given a target object whose pose evolves continuously, the robot must navigate in an unknown environment to approach and grasp it.
We follow the mainstream setup in presented \citep{zhang2024gamma, yokoyama2023asc, jauhri2022robot}.
The mobile manipulator consists of a planar base operating in $SE(2)$ with state $(x, y, \theta)$, an articulated arm with $n_a$ DoF, and a parallel-jaw gripper.
At each step $t$, the robot receives an RGB-D observation $(I, D)$ from a wrist-mounted camera and its proprioceptive state $\mathcal{S}_{\text{prop}}$, and outputs a whole-body action $(\mathcal{A}_{\text{base}}, \mathcal{A}_{\text{arm}}, \mathcal{A}_{\text{gripper}})$ for the base, arm and gripper.

\begin{figure}[t]
\centering
\includegraphics[trim=0.1cm 0.1cm 0cm 0.1cm, clip, width=0.95\linewidth]{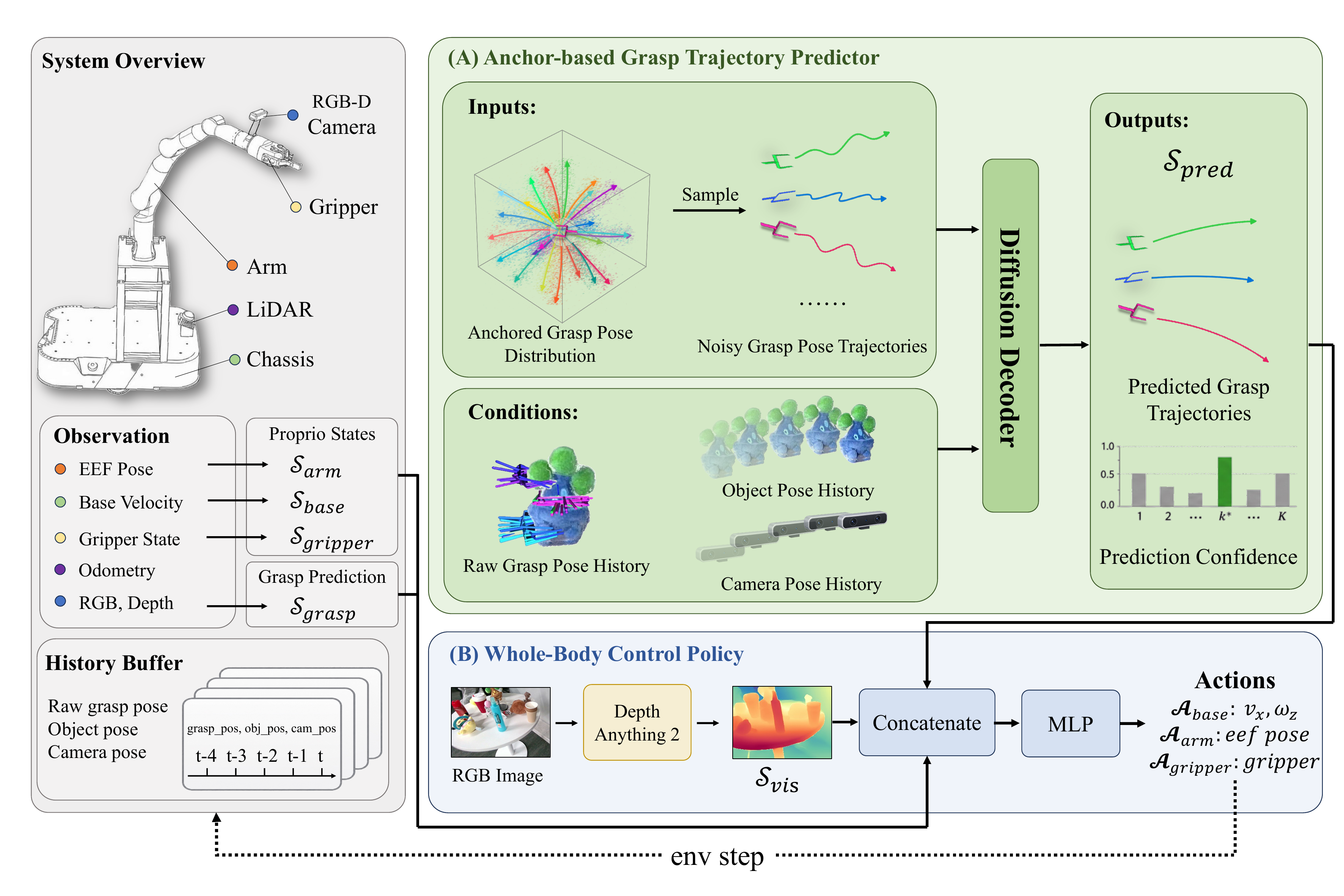}
\vspace{-10pt}
\caption{
\textbf{Overview of \ours{}.} 
Based on historical contexts, the anchor-based grasp trajectory predictor first generates candidate trajectories $\{\hat{\tau}_k\}_{k=1}^{K}$ with confidence scores $\mathbf{c}$. 
The highest-scoring trajectory is then selected and encoded with its score into a predictive feature $s_{\text{pred}}$. 
Finally, the whole-body policy integrates $\mathcal{S}_{\text{pred}}$, $\mathcal{S}_{\text{prop}}$, $\mathcal{S}_{\text{vis}}$, and $\mathcal{S}_{\text{grasp}}$ to output coordinated control actions $(\mathcal{A}_{\text{base}}, \mathcal{A}_{\text{arm}}, \mathcal{A}_{\text{gripper}})$.
}
\label{fig:pipeline}
\vspace{-15pt}
\end{figure}

\para{Overview.}
As shown in Fig.~\ref{fig:pipeline}, \ours{} consists of two modules.
(i)~An \emph{anchor-based grasp trajectory predictor} (Sec.~\ref{sec:predictor}) samples noisy grasp pose trajectories from a set of $K$ predefined anchors.
Conditioned on histories of camera and object poses, together with raw grasp candidates, the model denoises these trajectories into $K$ predicted grasp trajectories $\{\hat{\tau}_k\}_{k=1}^{K}$ with confidence scores $\mathbf{c} \in \mathbb{R}^K$, where each $\hat{\tau}_k = \{\hat{g}_{k,0}, \dots, \hat{g}_{k,T_{\text{seq}}}\}$ and $\hat{g}_{k,j} \in \mathbb{R}^9$ encodes position and 6D rotation.
(ii)~A \emph{whole-body control policy} $\pi_\theta$ (Sec.~\ref{sec:policy}) consumes the predictive feature $\mathcal{S}_{\text{pred}}$, proprioceptive state $\mathcal{S}_{\text{prop}}$, visual feature $\mathcal{S}_{\text{vis}}$ and fused grasp feature $\mathcal{S}_{\text{grasp}}$.
It outputs whole-body actions $(\mathcal{A}_{\text{base}}, \mathcal{A}_{\text{arm}}, \mathcal{A}_{\text{gripper}}) \in \mathbb{R}^{3+n_a}$.
The policy is trained with an anticipation-guided reward that adaptively blends the current observation and the predicted future grasp.


\subsection{Anchor-based Grasp Trajectory Predictor}
\label{sec:predictor}

Note that per-frame grasp poses from off-the-shelf detectors like GraspNet~\citep{fang2020graspnet} are designed for static scenes, which introduces two challenges in dynamic grasping.
First, these detectors cannot capture the multiple trajectory modes induced by diverse object motions.
Second, varying partial observations cause the top-scored grasp pose to shift across frames, yielding erratic trajectories.
To address these challenges, we introduce an anchor-based grasp trajectory predictor that maintains temporally consistent grasp poses while modeling this multi-mode trajectory structure.

\para{Anchor construction and truncated diffusion.}
We propose an anchor-based truncated diffusion model~\citep{liao2025diffusiondrive, wang2025trackvla} to efficiently capture this structure.
Instead of sampling from pure Gaussian noise, the model starts with $K$ predefined anchor trajectories and performs localized refinement.
To initialize these anchors, we apply K-Means to all training ground-truth grasp trajectories obtained from a sliding-window grasping fusion module~\citep{zhang2024gamma}, and use the resulting $K$ cluster centers as anchor trajectories, denoted as $\bar{\mathcal{T}}=\{\bar{\tau}_k|k=1,...,K\}$.
Each anchor is then perturbed via a truncated forward process that diffuses only up to a reduced horizon $T_{\text{trunc}}$:
\begin{equation}
\tilde{\tau}_k(t) = \sqrt{\bar{\alpha}_t}\,\bar{\tau}_k + \sqrt{1 - \bar{\alpha}_t}\,\epsilon, \qquad \epsilon \sim \mathcal{N}(0, I), \quad t \sim \mathcal{U}\{0,\dots,T_{\text{trunc}}-1\},
\label{eq:forward-diffusion}
\end{equation}
where $\{\bar{\alpha}_t\}$ follows a scaled-linear DDIM schedule.

\para{Conditional denoising.}
Our diffsion decoder $f_\theta$ takes the set of noisy anchor trajectories $\{\tilde{\tau}_k\}_{k=1}^{K}$ and a conditioning context $z$ as input,
and jointly outputs the denoised per-mode trajectories $\{\hat{\tau}_k\}_{k=1}^{K}$ together with their confidence scores $\mathbf{c} \in \mathbb{R}^{K}$:
\begin{equation}
\{\hat{\tau}_k, \hat{c}_k\}_{k=1}^{K} = f_\theta\!\left(\{\tilde{\tau}_k(t)\}_{k=1}^{K},\; z \right).
\label{eq:denoising}
\end{equation}
The conditioning context $z$ encodes three complementary cues.
First, a history of camera poses captures the egocentric motion induced by the robot's active viewpoint.
Second, a history of object poses captures the target motion over time.
Third, raw grasp pose candidates from GraspNet~\citep{fang2020graspnet} provide geometry-aware feasible grasp configurations.
Dedicated encoders separately map these three cues into $\mathbb{R}^d$ features, which the trajectory tokens query through cross-attention.
More architectural details are provided in the Appendix.

\para{Training.}
For each training sample with ground-truth trajectory $\tau_{\text{gt}}$, we assign the anchor trajectory closest to $\tau_{\text{gt}}$ as positive$\left(c_{k} = 1\right)$ and all the others as negative$\left(c_{k} = 0\right)$.
We then jointly optimize the trajectory regression loss and the score prediction loss. The loss is defined as:
\begin{equation}
\mathcal{L}
= \sum_{k=1}^{K}\left[ 
c_{k}\left|\tau_{\text{gt}} - \tau_{k} \right|_1  + \lambda \cdot BCE(c_k, \hat{c}_k)
\right]
\label{eq:training_loss}
\end{equation}
where $\lambda$ is a balancing weight.

\para{Inference.}
We use a truncated denoising process to obtain the final prediction. In detial, we initialize $K$ noisy trajectories by adding gaussian noise at truncation timestep $T_{trunc}$ to anchor trajectories $\bar{\mathcal{T}}$. A 2-step DDIM~\citep{song2020denoising} schedule is applied to predict clean trajectories $\{\hat{\tau}_k\}_{k=1}^{K}$ with conditioning context $z$. The trajectory with the highest confidence score is selected as the final output.

\subsection{Whole-Body Control Policy}
\label{sec:policy}

Given the predicted instantaneous grasp trajectory, the remaining challenge is to translate these features into effective feedforward whole-body control.
Naively relying on myopic reactive feedback often cause notable tracking lags during the final grasp.
When the robot is far from the object, simple tracking of the current-frame position is sufficient for a coarse approach. However, as it enters the imminent grasp window, our instantaneous grasp trajectory prediction becomes indispensable, providing the critical temporal prior to enable feedforward whole-body control before contact.
We therefore design a anticipation-guided reward that adaptively blends the two signals based on distance to the target, and train a whole-body control policy conditioned on the predictor's output.

\para{Policy formulation.}
The policy $\pi_\theta(\mathcal{A}_{\text{base}}, \mathcal{A}_{\text{arm}}, \mathcal{A}_{\text{gripper}} \mid \mathcal{S}_{\text{prop}}, \mathcal{S}_{\text{grasp}}, \mathcal{S}_{\text{vis}}, \mathcal{S}_{\text{pred}})$ takes as input a state $(\mathcal{S}_{\text{prop}}, \mathcal{S}_{\text{grasp}}, \mathcal{S}_{\text{vis}}, \mathcal{S}_{\text{pred}})$ composed of four parts.
$\mathcal{S}_{\text{prop}} = (P_{\text{ee}}, R_{\text{ee}}, \mathbf{v}_{\text{prop}})$ stacks the end-effector position $P_{\text{ee}}$, its 6D rotation $R_{\text{ee}}$, and a proprioceptive velocity vector $\mathbf{v}_{\text{prop}}$ of the base.
The feature vector $s_{\text{grasp}}$ aggregates grasp candidates generated by GraspNet~\citep{fang2020graspnet} within a local historical time window, thereby offering critical geometry-aware cues that enable the policy to generalize across objects of various shapes.
The visual feature $s_{\text{vis}}$ is derived from the Depth Anything V2 encoder~\citep{yang2024depth}, which captures the 3D structural characteristics and scene layout.
$s_{\text{pred}}$ contains the predicted grasp trajectory feature $\mathbf{f}_{k^\star}$ and the confidence scores $\mathbf{c}_{k}$ produced by the predictor of Sec.~\ref{sec:predictor}.
The action space is defined as $(\mathcal{A}_{\text{base}}, \mathcal{A}_{\text{arm}}, \mathcal{A}_{\text{gripper}})$, where $\mathcal{A}_{\text{base}}$ denotes the $2$-DoF differential-drive command for the mobile base, $\mathcal{A}_{\text{arm}}$ represents the residual end-effector pose of $6$-DoF for the arm, and $\mathcal{A}_{\text{gripper}}$ refers to $1$-DoF switch.

\para{Anticipation-guided reward.}
Let $g^{\text{cur}} \in \mathbb{R}^9$ be the current fused grasp pose. To adaptively balance long-range approach and short-range precision alignment, a distance-dependent trust coefficient $\alpha$ is computed based on the distance between the end-effector and the current grasp pose:
\begin{equation}
\alpha \,=\, \mathrm{clip}\!\left(\frac{\|P_{\text{ee}} - P_{\text{cur}}\| - d_{\text{near}}}{d_{\text{far}} - d_{\text{near}}},\ 0,\ 1\right),
\label{eq:alpha}
\end{equation}
where $P_{\text{ee}}$ and $P_{\text{cur}}$ denote the end-effector and current grasp positions, respectively. By design, $\alpha \to 1$ when the robot is far from the object, and $\alpha \to 0$ within the imminent grasp window.

The predicted instantaneous grasp trajectory $\hat{\tau} = \{\hat{g}_{0}, \dots, \hat{g}_{T_{\text{seq}}}\}$ is aggregated into a single future target $g^{\text{fut}}$ via $\alpha$-controlled softmax weights:
\begin{equation}
g^{\text{fut}} \,=\, \sum_{j=1}^{T_{\text{seq}}} w_j(\alpha)\, \hat{g}_j, \qquad
w(\alpha) \,=\, \mathrm{softmax}\!\left((1-\alpha) \cdot \mathbf{v}\right),
\label{eq:future-weight}
\end{equation}
where $\mathbf{v} = [T_{\text{seq}}, T_{\text{seq}}-1, \dots, 1]^T$ is a time-reversing logit vector. As the robot approaches the object, a declining $\alpha$ increases the value of $(1 - \alpha)$ toward 1. This mechanism adaptively sharpens the softmax distribution, scaling up the weight $w_1$ to dominate the aggregation. Because $\hat{g}_0$ represents the most immediate future step in the predicted trajectory, it provides the policy with the most reliable, localized guidance right before the grasp.
The final reward target $g^\star$ linearly interpolates between the aggregated future prediction and the current observation:
\begin{equation}
g^\star \,=\, (1 - \alpha)\, g^{\text{fut}} + \alpha\, g^{\text{cur}}.
\label{eq:target}
\end{equation}
When the robot is far ($\alpha \approx 1$), $g^\star$ approaches $g^{\text{cur}}$, guiding a stable, coarse approach via the real-time observation. Conversely, as the robot enters the tight grasp window ($\alpha \to 0$), $g^\star$ shifts entirely toward $g^{\text{fut}}$, leveraging our instantaneous grasp trajectory prediction to enable feedforward whole-body control before contact.

Finally, the future-aware reward $r_{\text{fut}}$ is defined to optimize both position and orientation alignment:
\begin{equation}
r_{\text{fut}} \,=\, \frac{1}{2}\!\left[\exp\!\left(-\frac{\|P^{\text{ee}} - P^\star\|}{\lambda_{\text{pos}}}\right) + \exp\!\left(-\frac{d_{\text{rot}}(R^{\text{ee}},\, R^\star)}{\lambda_{\text{rot}}}\right)\right],
\label{eq:r-fut}
\end{equation}
where $P^\star$ and $R^\star$ are the position and rotation components extracted from $g^\star$, $d_{\text{rot}}$ denotes the geodesic rotation distance, and $\lambda_{\text{pos}}$ and $\lambda_{\text{rot}}$ are control factors.

\para{Training.}
The policy is trained in three stages.
First, a teacher policy with access to privileged information is trained via reinforcement learning.
The teacher is then distilled into a student policy $\pi_\theta$ via DAgger~\citep{ross2011reduction}, where the student takes the full state $S$ defined above as input.
Finally, the student is fine-tuned via reinforcement learning with $r_{\text{fut}}$ as part of the reward.
The predictor kept frozen throughout. More training details are provided in the Appendix.


\section{Experiments}
\label{sec:experiments}

\subsection{Experimental Setup}

\para{Simulation environment.}
We evaluate all methods in Isaac Gym~\citep{makoviychuk2021isaac} with objects drawn from the YCB dataset~\citep{calli2015ycb}.
We design a dynamic object mobile manipulation (DOMM) benchmark, which comprises three settings that differ in the relative motion between the robot and the target, including Frontal interception, Lateral interception and Chasing.
At the start of each episode, a random YCB object is spawned within a setting-specific region with a randomized initial orientation and rotation angle. During each episode, the target object moves along a primary direction with small multi-directional perturbations.
In \emph{Frontal interception}, the primary trajectory is aligned along the robot's principal axis as the object moves toward it.
In \emph{Lateral interception}, the main direction spans across the robot's workspace from the side.
In \emph{Chasing}, the primary path extends outward as the robot pursues an object moving away.

\para{Baselines.}
To ensure a rigorous and fair evaluation under the intricate dynamics of DOMM, we evaluate all methods within a unified simulation infrastructure. Our comparative analysis focuses on baselines most relevant to our approach, categorized into reactive policies and prediction-augmented policies
For reactive baselines, \textbf{GAMMA}~\citep{zhang2024gamma} is a graspability-aware RL policy that fuses grasp pose candidates online to drive whole-body control.\textbf{ReachMM}~\citep{zhang2024gamma} is a variant of GAMMA that removes the online grasp-pose fusion module and instead conditions the policy solely on reachability states.
For prediction-augmented baselines, \textbf{\ours{}\textsubscript{KF}} and \textbf{\ours{}\textsubscript{Lin}} replace our anchor-based diffusion predictor with a Kalman filter and a linear-extrapolation module, respectively, while keeping the rest of the pipeline identical.

\para{Metrics.}
We evaluate each episode at the moment the grasp action is triggered and report three metrics.
\emph{Gaze Success Rate} (GazeSR) measures whether the wrist-mounted camera can observe the target object within a distance of 50\,cm at the time of grasping.
\emph{Grasp Success Rate} (GraspSR) measures whether the gripper's pose closely matches any densely annotated grasping pose, with deviations less than 10\,cm in distance and $10^{\circ}$ in angle.
\emph{Effective Step} (EStep) is the mean episode length across all trials, with failed episodes truncated to the maximum of 900 steps. 

\begin{table}[t]
\centering
\label{tab:modom-main}
\small
\setlength{\tabcolsep}{4pt}
\resizebox{\textwidth}{!}{%
\begin{tabular}{l ccc ccc ccc}
\toprule
& \multicolumn{3}{c}{\textbf{Frontal interception}} & \multicolumn{3}{c}{\textbf{Lateral interception}} & \multicolumn{3}{c}{\textbf{Chasing}} \\
\cmidrule(lr){2-4} \cmidrule(lr){5-7} \cmidrule(lr){8-10}
Method & GraspSR$\uparrow$ & GazeSR$\uparrow$ & EStep$\downarrow$ & GraspSR$\uparrow$ & GazeSR$\uparrow$ & EStep$\downarrow$ & GraspSR$\uparrow$ & GazeSR$\uparrow$ & EStep$\downarrow$ \\
\midrule
ReachMM~        & 3.5  & 53.0 & 871.5          & 2.0  & 54.0 & 885.8          & 6.0  & 58.0 & 855.3          \\
GAMMA~\citep{zhang2024gamma}          & 13.0 & 78.0 & \textbf{803.1}          & 6.5  & 63.0 & 853.9          & 4.5  & 44.5 & 864.4          \\
\midrule
\ours{}\textsubscript{KF}   & 0.5  & 87.0 & 896.1          & 0.5  & 80.0 & 895.7 & 1.5  & 80.5 & 890.4          \\
\ours{}\textsubscript{Lin}  & 3.0  & \textbf{90.0} & 885.9 & 3.0  & 77.0 & 883.1          & 3.0  & 83.5 & 875.5          \\
\ours{} (Ours)  & \textbf{39.0} & 44.0 & 826.8 & \textbf{62.5} & \textbf{80.5} & \textbf{647.0} & \textbf{91.5} & \textbf{97.5} & \textbf{301.7} \\
\bottomrule
\end{tabular}%
}
\vspace{2pt}
\caption{\textbf{Main results on DOMM across three motion settings.} Each setting underwent 200 trials. Best per column in \textbf{bold}.}
\label{tab:modom-main}
\vspace{-10pt}
\end{table}

\subsection{Results}

\para{Comparison with reactive baselines.}
As shown in Tab.~\ref{tab:modom-main}, \ours{} achieves the highest GraspSR across all three settings, outperforming the strongest reactive baseline by 26.0, 56.0, and 85.5 percentage points.
This gap mainly stems from the lack of prediction in reactive policies. They act on current observations, while \ours{} uses instantaneous grasp trajectories to guide the robot toward future contact poses. Since the object keeps moving during whole-body execution, a current-pose grasp can quickly become outdated. The predictor provides a short-horizon target, allowing the robot to intercept the object rather than merely follow the currently observed pose.

\para{Comparison with predictive variants.}
As shown in Tab.~\ref{tab:modom-main}, \ours{}\textsubscript{KF} and \ours{}\textsubscript{Lin} often maintain high GazeSR, but their GraspSR remains below 3.0\% in most settings.
This shows that adding a predictor is not sufficient by itself. A grasping pose is a geometry-conditioned target, and its instantaneous trajectory depends on both object motion and the evolving feasible contact configuration.
In our benchmark, objects exhibit substantial randomized motion, making such grasp trajectories difficult to model with simple Kalman filtering or linear extrapolation.
In contrast, our anchor-based diffusion predictor captures diverse grasp trajectory patterns and provides the policy with more reliable short-horizon guidance for whole-body control.

\begin{table}[t]
\centering
\label{tab:modom-ablation}
\small
\setlength{\tabcolsep}{4pt}
\resizebox{\textwidth}{!}{%
\begin{tabular}{l ccc ccc ccc}
\toprule
& \multicolumn{3}{c}{\textbf{Frontal interception}} & \multicolumn{3}{c}{\textbf{Lateral interception}} & \multicolumn{3}{c}{\textbf{Chasing}} \\
\cmidrule(lr){2-4} \cmidrule(lr){5-7} \cmidrule(lr){8-10}
Method & GraspSR$\uparrow$ & GazeSR$\uparrow$ & EStep$\downarrow$ & GraspSR$\uparrow$ & GazeSR$\uparrow$ & EStep$\downarrow$ & GraspSR$\uparrow$ & GazeSR$\uparrow$ & EStep$\downarrow$ \\
\midrule
w/o $S_{\text{grasp}}$  & 32.5  & 33.0 & 878.9 & 32.5  & 52.0 & 837.4 & 33.5  & 54.5 & 620.5\\
w/o $S_{\text{pred}}$   & 11.5  & \textbf{82.0} & \textbf{813.3} & 2.5  & 45.0 & 881.4 & 3.5 & 44.5 & 870.4\\
Vanilla Diffusion         & 32.0 & 34.5 & 896.0 & 49.0 & 78.0 & 776.4 & 42.0 & 76.5 & 765.1\\
\ours{} (Ours)  & \textbf{39.0} & 44.0 & 826.8 & \textbf{62.5} & \textbf{80.5} & \textbf{647.0} & \textbf{91.5} & \textbf{97.5} & \textbf{301.7} \\
\bottomrule
\end{tabular}%
}
\vspace{2pt}
\caption{\textbf{Ablation study on DOMM.} Each setting underwent 200 trials. Best per column in \textbf{bold}.}
\label{tab:modom-ablation}
\vspace{-20pt}
\end{table}

\subsection{Ablation Studies}
Tab.~\ref{tab:modom-ablation} verifies the contribution of each component.
Removing $S_{\text{grasp}}$ consistently lowers GraspSR, showing that grasp observations are important for dynamic grasping. They directly provide the policy with feasible contact cues to guide end-effector alignment.
Removing $S_{\text{pred}}$ causes a sharp drop in GraspSR, and the results even fall to a level comparable to the reactive baselines.
This confirms our insight that instantaneous grasp trajectory prediction are critical for dynamic grasping. Even when trained in dynamic environments, a policy without such predictive guidance struggles to achieve high GraspSR across diverse motion settings.
Replacing our anchor-based predictor with vanilla diffusion also reduces performance, indicating that anchor trajectories provide better priors for stabilizing multi-mode short-horizon prediction.
\subsection{Real-World Experiments}


\begin{figure}[t]
\centering
    \begin{minipage}[t]{0.52\linewidth}
        \vspace{0pt}
        \centering
        \includegraphics[width=0.9\linewidth]{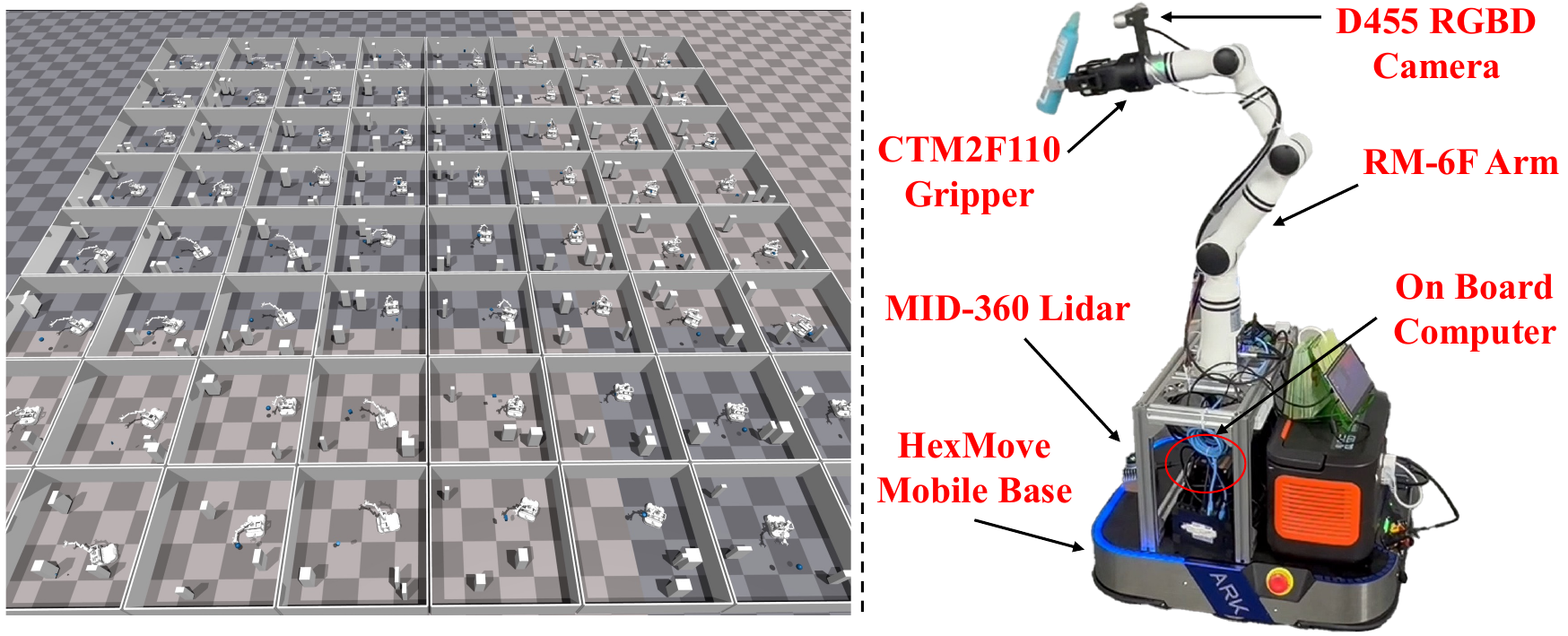}
        \vspace{-5pt}
        \captionof{figure}{\textbf{Experimental setup.} Left: parallel simulation environment in Isaac Gym. Right: real-world mobile manipulating system.}
        \label{fig:set_up}
    \end{minipage}%
    \hfill
    \begin{minipage}[t]{0.46\linewidth}
        \vspace{0pt}
        \centering

        {
        \renewcommand{\arraystretch}{1.1} 
        \setlength{\tabcolsep}{5pt}
        \setlength{\extrarowheight}{2pt}   

        \begin{tabular}{cc cc}
        \toprule
        \multicolumn{2}{c}{\textbf{Static}} & \multicolumn{2}{c}{\textbf{Dynamic}}\\
        \cmidrule(lr){1-2} \cmidrule(lr){3-4}
        Regular & Irregular & Easy & Hard  \\
        \midrule
        18 / 22 & 8 / 10 & 16 / 20 & 14 / 22 \\
        (81.9$\%$) & (80.0$\%$) & (80.0$\%$) & (63.6$\%$) \\
        \bottomrule
        \end{tabular}
        }

        \vspace{0.4em}
        \vspace{-5pt}
        \captionof{table}{\textbf{Sim-to-real experiments.} We report the success rate of \ours{} under both static and dynamic scenarios.}
        \label{tab:real-world}
    \end{minipage}
\vspace{-10pt}
\end{figure}

\begin{figure}[t]
\centering
\includegraphics[trim=0.1cm 0.1cm 0cm 0.1cm, clip, width=\linewidth]{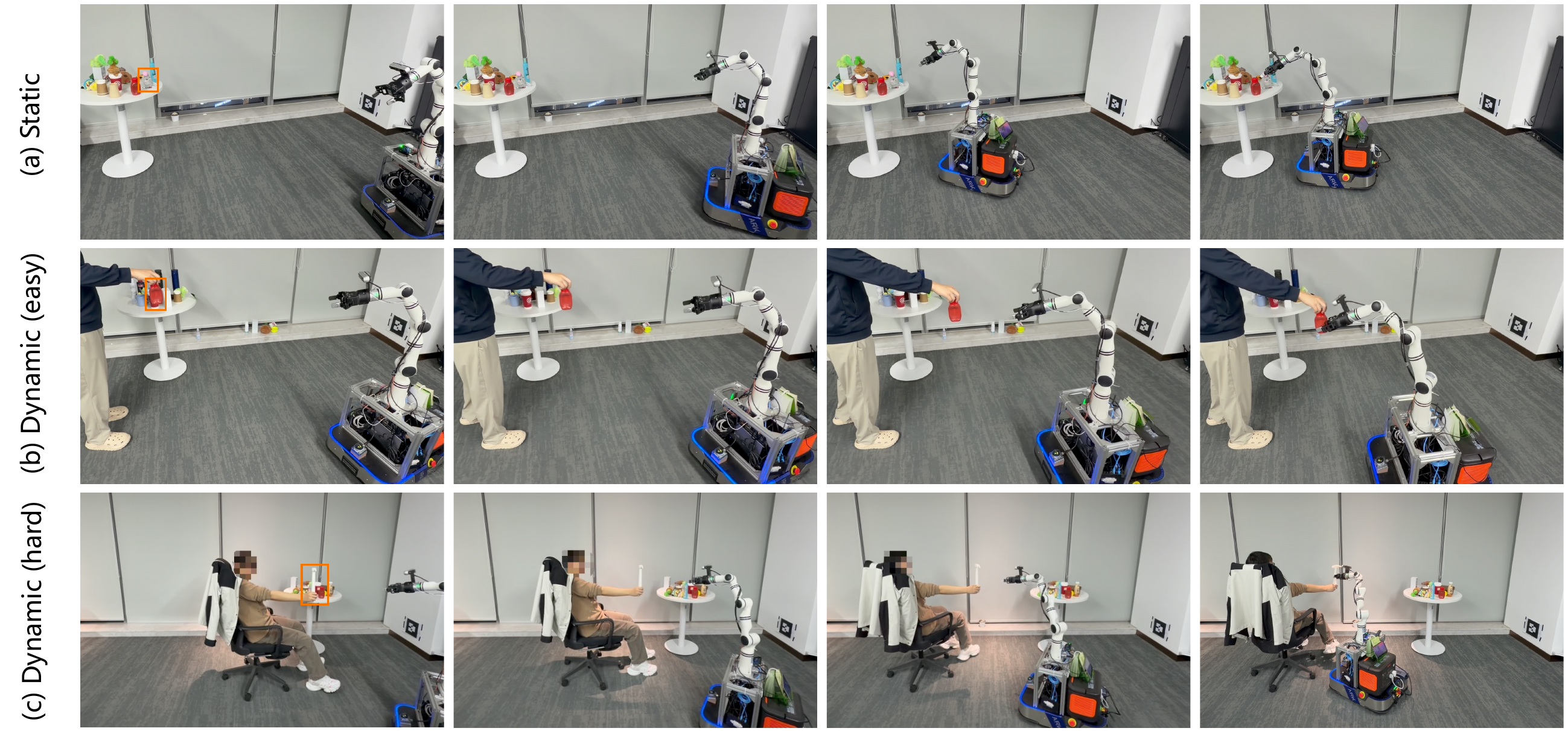}
\caption{
\textbf{Real-world qualitative results of \ours{}.} The target object is annotated by a bounding box in the first frame of each row.
}
\vspace{-10pt}
\label{fig:realworld}
\end{figure}

\para{Sim-to-real transfer.}
As shown in Fig.~\ref{fig:set_up}, We deploy \ours{} on a physical mobile manipulator comprising a Hexmove Echo differential-drive base and a 6-DoF RealMan RM65-6F arm~\citep{realman_rm65_6f}. For visual capture, an Intel RealSense D455 RGB-D camera is mounted on the gripper to provide egocentric observations. For localization, a base-mounted Livox MID-360 LiDAR runs Point-LIO~\citep{he2023point} to deliver real-time odometry and precise proprioceptive states. To detect target objects, we employ SAM2~\citep{ravi2025sam} to segment the egocentric point cloud, from which GraspNet~\citep{fang2020graspnet} extracts viable grasp poses. All the inputs are continuously fed into our policy to generate $9$-DoF actions. We conduct user study to demonstrate the practical robustness of our framework. Further implementation details are provided in the Appendix.

\para{User Study.} We evaluate our method on two different settings. For \emph{static scenes}, the objects are grouped into regular and irregular categories based on their shapes. Regular objects include bottles and boxes, while irregular objects are presented by plush toys. All objects are randomly replaced on a clusttered tabeltop. For \emph{dynamic scenes}, the tasks are divided into easy and hard modes based on human interaction profiles. In the easy mode, the user seamlessly hands over the object to the robot. In the hard mode, the user actively exhibits adversarial actions, such as moving the object away or swaying it side-to-side. The experimental results are shown in Tab.~\ref{tab:real-world} and qualitative visualizations are displayed in Fig.~\ref{fig:realworld}. The data indicates that our approach achieves a high success rate under varying real-world deployment conditions.


\section{Conclusion}
\label{sec:conclusion}

We present \ours{}, a dynamic mobile grasping framework that couples instantaneous grasp trajectory prediction with a whole-body control policy.
At its core, an anchor-based diffusion predictor refines predefined grasp trajectory anchors through truncated denoising, producing temporally consistent, multi-mode grasp trajectories conditioned on historical observations.
The resulting predictive features are integrated into a whole-body control policy through an anticipation-guided reward, which adaptively balances current observations and predicted instantaneous grasp poses to support stable approach from afar and feedforward control within the imminent grasp window.
Experiments on the DOMM benchmark show that \ours{} consistently outperforms reactive and prediction-augmented baselines, while real-world deployment further demonstrates robust dynamic grasping under diverse object motions.

\para{Limitations.}
Despite its effectiveness, our method still has several limitations.
First, it does not yet address long-horizon dynamic grasping tasks.
Second, the current framework may fail when the target undergoes overly aggressive motion.
Finally, the whole-body control policy can be less reliable in narrow regions or constrained passages, where limited free space makes coordinated base-arm motion difficult.


\clearpage


\bibliography{example}  

\end{document}